\documentclass[]{uiuc_paper}
\pdfoutput=1

\usepackage[toc,page,header]{appendix}
\usepackage{minitoc}
\usepackage{cleveref}
\usepackage{subcaption}
\usepackage{booktabs}
\usepackage{tabularx}
\usepackage{colortbl}
\usepackage{algorithm}
\usepackage{algpseudocode}
\usepackage{inconsolata}
\usepackage{needspace}

\usepackage{latexsym}
\usepackage{url}
\usepackage{amssymb}
\usepackage{amsmath}
\usepackage{amsfonts}
\usepackage[utf8]{inputenc}
\usepackage{booktabs}
\usepackage{pifont}
\usepackage{multirow}
\usepackage{makecell}
\usepackage{paralist}
\usepackage{xspace}
\usepackage{adjustbox}

\newcommand{\method}{TF-OPD\xspace}

\definecolor{cnavy}{HTML}{2F4A6D}
\definecolor{cbluedeep}{HTML}{506684}
\definecolor{cbluemid}{HTML}{8797B0}
\definecolor{cbluesoft}{HTML}{B7C6DC}
\definecolor{cice}{HTML}{DCE2EA}
\definecolor{cpink}{HTML}{DCC2BE}
\definecolor{cink}{HTML}{2B3F5B}
\definecolor{crose}{HTML}{A47C7A}
\colorlet{oursrow}{cbluesoft!28}

\newcommand{\dd}{\Delta}
\newcommand{\Gt}{\widetilde{G}}
\newcommand{\Wb}{W_{\mathrm{b}}}
\newcommand{\tr}{\operatorname{tr}}

\title{Training-Free On-Policy Distillation}

\author[1,*]{Bangji Yang}
\author[1,*]{Jiajun Fan}
\author[1]{Zhaoyu Zhu}
\author[2]{Hongbo Ma}
\author[1,\textdagger]{Ge Liu}

\affiliation[1]{University of Illinois at Urbana-Champaign}
\affiliation[2]{Tsinghua University}

\contribution[*]{Equal contribution}
\contribution[\textdagger]{Corresponding author}

\abstract{On-policy distillation consolidates specialist models into one student at the
price of rollouts, teacher passes and gradient steps for every teacher. We show
that the consolidation does not need training: with forward passes only, the
capabilities of separately tuned specialists can be aggregated into one model
that is, on average, more accurate than either specialist alone. When
specialists share a base model, distillation has an exact layer-wise solution:
each linear layer of the fused model must reproduce every specialist's output on
the inputs that specialist receives while reading its own answers, a
least-squares problem whose solution needs only the second moments of those
inputs. One forward pass over a few hundred rollouts provides them. We call this
training-free on-policy distillation (\method). What the rollouts contain
matters more than the solver: answers truncated in repetition loops are few
among a mathematics specialist's rollouts but carry a large share of their
tokens, and removing them lifts the fused model on AIME25 from task arithmetic's
level to the specialist's at matched precision. On Olmo~3 7B, a few GPU-hours of
forward computation per specialist yield a generalist that keeps most of each
specialist's gain on each of seven mathematics and code benchmarks.
}

\date{\today}
\correspondence{\email{geliu@illinois.edu}}

\begin{document}

\newtheorem{theorem}{Theorem}[section]
\newtheorem{proposition}[theorem]{Proposition}
\newtheorem{lemma}[theorem]{Lemma}
\newtheorem{corollary}[theorem]{Corollary}
\newtheorem{definition}[theorem]{Definition}
\newtheorem{assumption}[theorem]{Assumption}
\newtheorem{remark}[theorem]{Remark}

\maketitle

\section{Introduction}
\label{sec:intro}

Reinforcement learning from verifiable rewards has become the standard way to
turn a base language model into a strong reasoner in one domain
\citep{shao2024deepseekmath,deepseek2025r1,yu2025dapo,he2025justrl,luo2025deepscaler}.
Because the reward is domain-specific, the result is a family of specialists
that share an initialization: from one base checkpoint, one model is trained
on competition mathematics, another on programming, a third on instruction
following \citep{olmo2025olmo3}. Each is far ahead on its own domain and far
behind on the others, so a user who wants mathematics and programming in the
same model has to build that model from the specialists.

\begin{figure}[t]
\centering
\includegraphics[width=\textwidth]{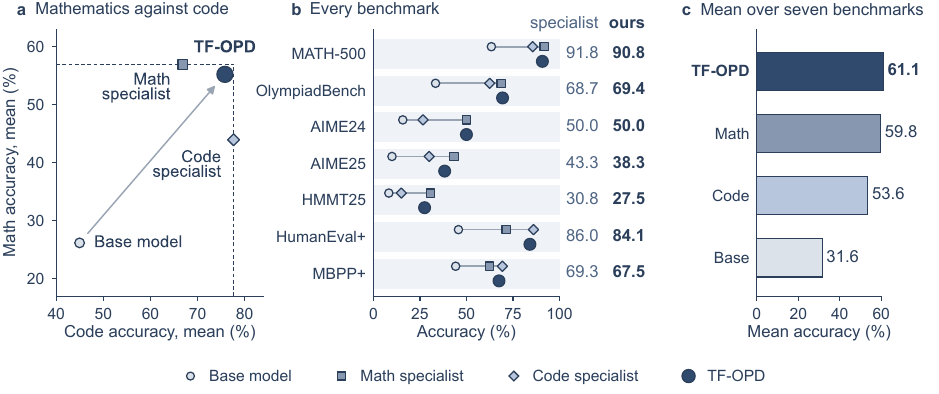}
\caption{\textbf{A generalist with the highest mean accuracy, without
training (Olmo~3 7B).} (a)~Mean accuracy on the five mathematics benchmarks
against mean accuracy on the two code benchmarks; dashed lines mark what each
specialist reaches on its own domain, and \method, computed from the two with
forward passes only, lies next to the corner they define. (b)~Every
benchmark: the base model and the specialists on the upper line of each row,
\method below, and at the right the score of that domain's specialist and of
\method. (c)~Mean accuracy over the seven benchmarks: \method is more accurate
on average than either specialist it was built from.}
\label{fig:teaser}
\end{figure}

The available routes either train or ignore the rollouts
(Table~\ref{tab:routes}). One can repeat reinforcement learning on the union
of the domains \citep{liu2025prorl,liang2026boosting}, or distill: on-policy
distillation lets the student generate, has a teacher supervise those
generations, and updates the student on its own distribution
\citep{ross2011dagger,agarwal2024gkd,gu2024minillm}. It has become a
post-training route for consolidating specialists into one student
\citep{hou2026uniopd}, at the price of rollouts, teacher passes and gradient
steps for every teacher. The training-free alternative is to merge weights.
Arithmetic on parameters \citep{ilharco2023task} does not know which inputs
the parameters act on, and on reinforcement-learned specialists it can give
up part of what made them specialists: summing the mathematics and coding
updates of a 7B model leaves it below the mathematics specialist on AIME25
(Section~\ref{sec:exp-selection}). RegMean \citep{jin2023regmean} does use
inputs, in a closed-form least-squares solution, but takes them from each
task's training corpus, whereas a reasoning model spends almost all of its
computation on text it has generated itself.

\begin{table}[t]
\centering\small
\setlength{\tabcolsep}{8pt}
\begin{tabular}{@{}lccc@{}}
\toprule
Route & No training & On-policy inputs & Matches teachers \\
\midrule
Multi-domain RL             & $\times$     & $\checkmark$ & $\times$ \\
On-policy distillation      & $\times$     & $\checkmark$ & $\checkmark$ \\
Task arithmetic, TIES, DARE & $\checkmark$ & $\times$     & $\times$ \\
RegMean                     & $\checkmark$ & $\times$     & $\checkmark$ \\
\rowcolor{oursrow}[\tabcolsep][\tabcolsep]\method (ours) & $\checkmark$ & $\checkmark$ & $\checkmark$ \\
\bottomrule
\end{tabular}
\caption{\textbf{Routes from several specialists to one model.} \method needs
no training, gathers its statistics on text the models generate, and is fit
to match the specialists' outputs.}
\label{tab:routes}
\end{table}

We connect the two lines. For a single linear layer, asking the fused weight
to reproduce each teacher's output on that teacher's inputs is the
distillation objective restricted to the layer, and it is the least-squares
problem that RegMean solves: once the second moments of the inputs are known,
no optimization is left to run. \method gathers those moments where on-policy
distillation gathers its data, on rollouts. Each specialist answers a few
hundred training problems from its domain, the answers are filtered, the
specialist reads its kept answers once with teacher forcing while forward
hooks accumulate a Gram matrix per layer, and each layer is then solved
exactly. Nothing is back-propagated.

On Olmo~3 7B this yields a generalist without a single gradient step: one
model that keeps most of what each specialist gained on five mathematics and
two code benchmarks and that is, on average, more accurate than either
specialist alone (Figure~\ref{fig:teaser}).
Three findings explain when it
works. First, the filter carries the result. A minority of the mathematics
specialist's rollouts run to the length limit in a repetition loop: few among
the answers, several times more among the tokens. With them the closed form is
no better than task arithmetic on AIME25, and removing them alone closes the
gap to the specialist scored in the same format. Second, the token-level fit is
not what fails: the unfiltered model already follows the specialist almost
token for token, so hard problems are decided by which inputs the statistics
weigh. Third, rollouts written by the student, the on-policy data of trained
distillation, do not improve the closed form. Reinforcement learning also
moves weights so little that the 16-bit release format distorts a fused
update, which makes the formats of storage and evaluation part of the method.

\begin{figure}[t]
\centering
\includegraphics[width=\textwidth]{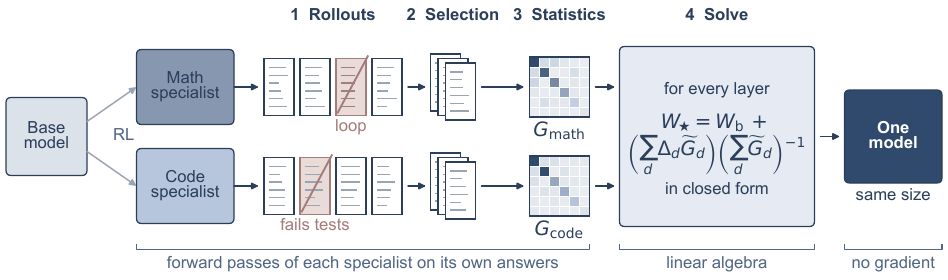}
\caption{\textbf{Overview of \method.} Two specialists trained from one base
model each answer training problems of their own domain~(1); answers that end
in a repetition loop or fail grading are dropped~(2); each specialist reads its
kept answers once while hooks record the second moments $G_d$ of every layer's
inputs~(3); and each layer of the fused model is obtained by one linear
solve~(4). Steps 1 to 3 are forward passes and step 4 is linear algebra, so no
gradient is computed at any point. Step 2 decides the hard problems
(Section~\ref{sec:selection}).}
\label{fig:overview}
\end{figure}

\needspace{6\baselineskip}
\paragraph{Contributions.}
\begin{itemize}
\item We show that the capabilities of separately tuned specialists can be
aggregated without training. On Olmo~3 7B, a closed-form fusion of a
mathematics and a coding specialist keeps most of each specialist's gain on
seven benchmarks and is, on average, more accurate than either specialist
alone, after a few GPU-hours of forward computation per specialist.
\item The method, \method, reads layer-wise distillation from several
teachers as the closed-form problem of RegMean and feeds it statistics from
the specialists' own rollouts, using forward passes only.
\item We show that the composition of those rollouts decides the outcome on
hard problems. Truncated answers ending in repetition loops are a small share
of answers and a large share of tokens, and removing them separates a fused
model that falls short of its mathematics specialist from one that matches it
in the same evaluation format.
\item We analyze what matters and what does not: a close token-level fit does
not protect hard problems; student rollouts, domain reweighting and mined
hard problems do not improve on selection; numerical precision does.
\end{itemize}

\section{Related Work}
\label{sec:related}

\paragraph{Knowledge distillation.} Classical distillation fits a student to a
teacher's output distribution on a fixed corpus \citep{hinton2015distilling}
or to sequences the teacher generates \citep{kim2016sequence}. On-policy
formulations supervise text generated by the model being trained, following
the imitation-learning argument for learning on the learner's own state
distribution \citep{ross2011dagger,agarwal2024gkd,gu2024minillm}; the idea has
been applied to reasoning models, including with a model as its own teacher
\citep{zhao2026opsd}, and knowledge fusion extends distillation to several
source models \citep{wan2024fusellm}. Recent work asks which rollouts
on-policy distillation should use, selecting a few hard problems
\citep{hou2026whatmatters}, weighting problems by the student's pass rate
\citep{xu2026paced}, or balancing student-generated states across teachers
\citep{hou2026uniopd}. All of these are optimization procedures. \method keeps
two of their ingredients, a match-the-teacher objective and statistics
gathered on model-generated text, and removes the optimization by solving the
objective layer by layer. Our finding that the composition of the rollouts
decides the outcome is a closed-form counterpart of theirs.

\paragraph{Model merging.} Weight averaging across fine-tuning runs improves
accuracy at no inference cost \citep{wortsman2022soups}; task arithmetic
composes updates relative to a common initialization \citep{ilharco2023task},
and TIES-Merging and DARE reduce interference among summed updates
\citep{yadav2023ties,yu2024dare}. Spectral methods exploit the matrix structure
of updates \citep{gargiulo2024tsv,marczak2025isotropic,sun2026resmerge}. These
methods read weights only. Weight-space edits also target reasoning models,
to amplify one reinforcement-learning run
\citep{yang2026doesscalingreinforcementlearning} or to make a model more
efficient \citep{ma2026s3spectralnullspaceswap}; we combine runs on different
domains. Fisher merging and RegMean bring in data through parameter
importance and through input second moments respectively
\citep{matena2021fisher,jin2023regmean}. \method uses RegMean's closed form;
what differs is what it is fed and what it is used for. RegMean was introduced
for models fine-tuned on supervised tasks, with statistics computed on each
task's training inputs. The specialists here are reinforcement-learned
reasoners whose inputs are overwhelmingly their own long generations, so the
statistics have to come from rollouts, and we show that on AIME25 unfiltered
rollouts gain nothing over task arithmetic, whereas filtered ones recover the
specialist.

\paragraph{Other training and evaluation settings.} Multi-domain fine-tuning
models interactions among samples during training \citep{liang2025boosting};
distributional Pareto-optimal reinforcement learning studies trade-offs across
objectives \citep{NEURIPS2023_32285dd1}. Policy optimization has also been
studied with interactive guidance
\citep{liu2026socraticpopolicyoptimizationinteractive}, while anomaly-guided
learning addresses imperfect demonstrations \citep{AAMAS'22:AGPO}. Related
choices about supervision include adaptive sample scheduling for direct
preference optimization \citep{huang2025adaptive} and reward models that go
beyond semantics \citep{Huang2026RealTimeAR}. Arena-based LLM evaluation
\citep{liu2025elo} and medical vision-language work on adaptive causal
reasoning \citep{lin2026models} and latent-memory evolution
\citep{zhu2026medsynapsevbridgingvisualperception} concern other evaluation or
application settings.

\section{Method}
\label{sec:method}

\subsection{Distillation of one layer is least squares}
\label{sec:objective}

Let $\theta_{\mathrm{b}}$ be a base model and $\theta_1,\dots,\theta_K$
specialists post-trained from it, one per domain; we call them teachers, and
the model we construct, $\theta_\star$, the student. All share an architecture.
Consider a linear layer with base weight $\Wb\in\mathbb{R}^{m\times n}$ and
teacher weights $W_d=\Wb+\dd_d$. Let $\mu_d$ be the distribution of the input
$x\in\mathbb{R}^n$ this layer receives inside teacher $d$, to be specified in
Section~\ref{sec:stats}. We ask the student's layer to reproduce each teacher's
output on that teacher's inputs,
\begin{equation}
W_\star=\arg\min_{W}\ \sum_{d=1}^{K} w_d\,\mathbb{E}_{x\sim\mu_d}\big\|Wx-W_dx\big\|_2^2 ,
\label{eq:obj}
\end{equation}
with domain weights $w_d>0$. This is the distillation objective restricted to
one layer, with the teacher's activations as inputs. It is quadratic in $W$
and depends on $\mu_d$ only through the second moment
$G_d=\mathbb{E}_{x\sim\mu_d}[xx^{\!\top}]$. Setting the gradient to zero gives
$W_\star\sum_d w_dG_d=\sum_d w_dW_dG_d$, and since the base weight cancels,
\begin{equation}
W_\star=\Wb+\Big(\sum_{d}\dd_d\,\Gt_d\Big)\Big(\sum_{d}\Gt_d\Big)^{-1},
\label{eq:solve}
\end{equation}
where $\Gt_d=w_dG_d$. This is the RegMean solution \citep{jin2023regmean}
written for updates. Following that work we shrink off-diagonal second moments,
\begin{equation}
\Gt_d=w_d\big[\alpha\,G_d+(1-\alpha)\,\mathrm{diag}(G_d)\big],\qquad \alpha=0.9,
\label{eq:shrink}
\end{equation}
which makes the system well conditioned and is equivalent to adding a
per-coordinate penalty that pulls $W$ toward each teacher
(Appendix~\ref{app:derivation}).

Equation~\eqref{eq:solve} makes the contrast with weight arithmetic explicit.
Task arithmetic sets the update to $\sum_d\dd_d$ and so applies every
teacher's update to every input, including inputs that only another teacher
ever sees; on teacher $e$'s inputs the layer output is off by
$\sum_{d\ne e}\dd_dx$. \method minimizes exactly that error, weighted by how
often each input direction occurs. Three properties of the solution locate
where teachers can conflict.

\begin{proposition}
\label{prop:conflict}
Let $\dd_\star=W_\star-\Wb$ solve Equation~\eqref{eq:solve} with
$\Gt_d=w_dG_d$.
\begin{enumerate}
\item[(i)] If $G_d=G$ is invertible for all $d$, then
$\dd_\star=\sum_dw_d\dd_d/\sum_dw_d$.
\item[(ii)] If $G_e$ is invertible and $w_d/w_e\to0$ for all $d\ne e$, then
$\dd_\star\to\dd_e$.
\item[(iii)] If the ranges of $G_1,\dots,G_K$ are mutually orthogonal and the
inverse is taken as the pseudo-inverse, then $\dd_\star x=\dd_ex$ for every
$x$ in the range of $G_e$ and every $e$.
\end{enumerate}
\end{proposition}

Appendix~\ref{app:derivation} gives the proofs. By (i), the closed form reduces
to weight averaging, not to task arithmetic, when the teachers cannot be told
apart by their inputs. By (iii), interference arises only where input
distributions overlap; Equation~\eqref{eq:diag} below measures how much.

\subsection{On-policy statistics}
\label{sec:stats}

The solution is only as good as $\mu_d$. We take $\mu_d$ to be the distribution
of layer inputs when teacher $d$ processes its own answers to problems from
its domain. This is the distribution the layer operates on in use, since a
reasoning model's context is dominated by its own generation, and it cannot be
obtained from reference solutions, which the model did not write. We refer to
statistics gathered on model-generated rollouts as on-policy, in contrast to
statistics gathered on a fixed corpus; the rollouts in our main configuration
are the teachers', as in sequence-level distillation \citep{kim2016sequence},
and Section~\ref{sec:ablations} evaluates the variant that uses the student's.

Concretely, teacher $d$ answers a set of training problems from its domain,
sampled exactly as in evaluation. After selection
(Section~\ref{sec:selection}), the teacher reads each kept sequence, prompt
and answer, in a single teacher-forced forward pass. A forward pre-hook on
every linear layer of every decoder block accumulates $X^{\!\top}X$ over all
token positions, and $G_d$ is that sum divided by the number of tokens, so
that domains with longer answers do not dominate by length. Layers that read
the same tensor share a Gram matrix: in the architectures we use, the query,
key and value projections form one group, the gate and up projections another,
and the attention output and down projections one each. The grouping is
detected automatically from one short forward pass, and the same statistics
are collected for the final hidden states to solve the output head.

\subsection{Solving}
\label{sec:solve}

For each group, Equation~\eqref{eq:solve} is solved once per member layer in
double precision. Tensors that are not linear maps of a hidden state ---
embeddings, normalization gains and biases --- change little and are combined
by task arithmetic. Besides the fused model, the solve returns for every layer
and teacher the share of that teacher's update left unreproduced,
\begin{equation}
\varepsilon_d=\frac{\tr\!\big[(\dd_\star-\dd_d)\,\Gt_d\,(\dd_\star-\dd_d)^{\!\top}\big]}
{\tr\!\big[\dd_d\,\Gt_d\,\dd_d^{\!\top}\big]},
\label{eq:diag}
\end{equation}
with $\Gt_d$ of Equation~\eqref{eq:shrink}. It is zero when the fused layer
matches the teacher on its inputs, and it locates the layers where the
teachers compete. Figure~\ref{fig:overview} illustrates the procedure, and
Algorithm~\ref{alg:main} in Appendix~\ref{app:eval} states it in full.

\subsection{Which rollouts to use}
\label{sec:selection}

Every token of every kept answer contributes one outer product to $G_d$, so
the statistics are a token-weighted picture of the teacher's behavior. That
picture is distorted by a failure mode of long-form generation. Some answers
do not terminate: the model enters a repetition loop and continues to the
length limit. Such answers are few, but each is as long as the limit allows,
and the layer inputs inside a loop are nearly periodic, so their contribution
to the Gram matrix is large and concentrated on a few directions. The
least-squares solution then spends its capacity reproducing the teacher on
degenerate input.

We detect loops without supervision from the compressibility of the text: an
answer is flagged when the last 20{,}000 characters compress, under DEFLATE, to
less than 5\% of their length, far below ordinary reasoning text. Our default
selection keeps answers that finished on their own and are not flagged. Where
answers can be graded automatically, we additionally keep only correct ones:
mathematics answers are checked against the reference answer, and code
answers are run against the tests of their problem.

\subsection{Precision}
\label{sec:precision}

Released checkpoints are stored in bfloat16, which keeps eight significant
bits. Reinforcement learning with small learning rates barely moves weights at
that resolution, by about one representable step
(Section~\ref{sec:exp-precision}). A fused update is no larger, and unlike a
specialist's it is not made of whole steps. We therefore store fused models
in 32-bit floats, since rounding the fused update to bfloat16 measurably
distorts it. A fused model is evaluated in float16, which keeps three more
significant bits than bfloat16 and fits the memory of a single GPU where
float32 does not. A format change by itself shifts measured accuracy, so we
state the format behind every score.

\section{Experiments}
\label{sec:exp}

\subsection{Setup}
\label{sec:setup}

\paragraph{Models.} The main experiments use the Olmo~3 RL-Zero family
\citep{olmo2025olmo3}. The base is Olmo-3-1025-7B, and the teachers are the
final checkpoints of Olmo-3.1-7B-RL-Zero-Math and Olmo-3.1-7B-RL-Zero-Code,
each obtained by reinforcement learning from that base on a single domain.

\paragraph{Rollouts and statistics.} For mathematics the teacher answers 256
training problems from DAPO-Math-17k \citep{yu2025dapo}; for code, 256
training problems from the DeepCoder training set \citep{luo2025deepcoder},
with the instruction used by EvalPlus. Each problem is answered once.
Mathematics is sampled at temperature 0.7 and top-$p$ 0.9 with a limit of
31{,}744 new tokens, code at temperature 1.0 and top-$p$ 1.0 with a limit of
32{,}768, matching evaluation. No evaluation problem is used. Selection keeps
the 175 mathematics answers that are correct, finished and loop-free, and the
92 code answers that pass every test of their problem. We use $\alpha=0.9$
and equal domain weights.

\paragraph{Evaluation.} Mathematics: AIME24, AIME25 and HMMT25, the latter
two from MathArena \citep{balunovic2025matharena}, with accuracy averaged
over four sampled answers per problem and graded by rule, rejected answers
being re-checked by a model verifier (CompassVerifier-3B); and the two larger
benchmarks, MATH-500 \citep{hendrycks2021math} and OlympiadBench
\citep{he2024olympiadbench}, with one sampled answer per problem. Code:
HumanEval+ and MBPP+ under EvalPlus
\citep{chen2021codex,austin2021mbpp,liu2023evalplus}, with one sampled answer
per problem. Released checkpoints are evaluated in bfloat16, the format they
are released in, and fused models in float16 (Section~\ref{sec:precision});
Appendix~\ref{app:eval} gives the full protocol.

\paragraph{Baselines.} The base model and each teacher. Task arithmetic with
unit coefficients, $\Wb+\dd_{\text{math}}+\dd_{\text{code}}$
\citep{ilharco2023task}, is the weight-only reference;
Section~\ref{sec:ablations} adds a low-rank variant of it and closed-form
variants that differ in the rollouts they use.

\begin{table}[t]
\centering\small
\setlength{\tabcolsep}{4.8pt}
\begin{tabular}{@{}lcccccccc@{}}
\toprule
 & \multicolumn{5}{c}{Mathematics} & \multicolumn{2}{c}{Code} & \\
\cmidrule(lr){2-6}\cmidrule(lr){7-8}
Model & MATH-500 & OlympiadBench & AIME24 & AIME25 & HMMT25 & HumanEval+ & MBPP+ & Mean \\
\midrule
Base model & 63.4 & 33.4 & 15.8 & 10.0 & 8.3 & 45.7 & 44.2 & 31.6 \\
Math specialist & \textbf{91.8} & 68.7 & \textbf{50.0} & \textbf{43.3} & \textbf{30.8} & 71.3 & 62.4 & 59.8 \\
Code specialist & 85.6 & 62.5 & 26.7 & 30.0 & 15.0 & \textbf{86.0} & \textbf{69.3} & 53.6 \\
\rowcolor{oursrow}[\tabcolsep][\tabcolsep]\method (ours) & 90.8 & \textbf{69.4} & \textbf{50.0} & 38.3 & 27.5 & 84.1 & 67.5 & \textbf{61.1} \\
\bottomrule
\end{tabular}
\caption{\textbf{One generalist, ahead of both specialists on average (Olmo~3 7B).}
Accuracy (\%). AIME24, AIME25 and HMMT25 average four sampled answers per
problem; MATH-500, OlympiadBench, HumanEval+ and MBPP+ use one. Released
checkpoints are evaluated in bfloat16 and the fused model in float16. The mean
is taken over the seven benchmarks before rounding. The best value of each
column is in bold.}
\label{tab:main}
\end{table}

\subsection{Main results}
\label{sec:main}

Figure~\ref{fig:teaser} and Table~\ref{tab:main} give the main comparison.
The fused model has the highest mean accuracy of the four models
(Figure~\ref{fig:teaser}c): built without a gradient step, it is on average
better than either specialist it was built from. The reason is a trade-off
that each specialist carries and the fused model does not. Each specialist
is far ahead on its own domain and far behind on the other; \method is close
to the specialist of the domain on every benchmark
(Figure~\ref{fig:teaser}b), and in the plane of mathematics against code it
sits beside the corner set by the two specialists' own-domain scores
(Figure~\ref{fig:teaser}a). It equals the mathematics specialist on AIME24
and is slightly ahead of it on OlympiadBench, is nearly level with the
specialist of the domain on MATH-500 and on both code benchmarks, and keeps
most of what reinforcement learning added on every benchmark. It is the only
model that is never far from the best score, and its lead in mean accuracy
does not come from the benchmark count:
weighting the two domains equally widens it (Figure~\ref{fig:teaser}a).

None of this required training. The procedure is one generation pass and one
teacher-forced pass per specialist, followed by a linear solve per layer. It
keeps no optimizer state, has no learning rate to tune, and finishes in a few
GPU-hours per specialist (Table~\ref{tab:cost} in Appendix~\ref{app:eval}).
Training-based on-policy distillation spends the same kind of forward
computation at every update and adds a backward pass each time. For
specialists that share a base model, the closed form reaches a consolidated
model at roughly the cost of collecting such data once. Its advantage over
weight arithmetic is on hard problems: task arithmetic costs even less, but
falls short on AIME25, where the closed form with clean rollouts matches the
specialist in the same format (Section~\ref{sec:ablations}).

On AIME25 the fused model is within a point of the mathematics specialist
once both are scored in float16 (Table~\ref{tab:abl-aime});
Table~\ref{tab:main} scores the released checkpoints in their native format.
The three competition benchmarks have thirty problems each, so differences
of a few points are within what they resolve.

\subsection{Case study}
\label{sec:cases}

\begin{figure}[t]
\centering\footnotesize
\setlength{\fboxsep}{4pt}\setlength{\fboxrule}{0.7pt}
\fcolorbox{cink}{cice}{\parbox{\dimexpr\textwidth-2\fboxsep-2\fboxrule}{%
\textbf{Problem (HMMT25).} In a two-dimensional cave with a parallel floor
and ceiling, two stalactites of lengths 16 and 36 hang perpendicularly from
the ceiling, while two stalagmites of heights 25 and 49 grow perpendicularly
from the ground. If the tips of these four structures form the vertices of a
square in some order, compute the height of the cave. \hfill
\textbf{Answer: 63}}}

\vspace{4pt}
\begin{minipage}[t]{0.492\textwidth}
\fcolorbox{cnavy}{white}{\parbox[t]{\dimexpr\linewidth-2\fboxsep-2\fboxrule}{%
\textbf{\textcolor{cnavy}{\method: answers 63, correct}} \hfill 3 of 4 samples\\[3pt]
\emph{``Suppose that the four points form a square. Therefore, the midpoints
of the diagonals must coincide.''}\\[1pt]
\hspace*{1em}$\downarrow$\\[1pt]
\emph{``Midpoint of AB: (\,\dots, $H-26$). Midpoint of CD: (\,\dots, 37).''}\\[1pt]
\hspace*{1em}$\downarrow$\\[1pt]
\emph{``\dots\ their y-coordinates must satisfy $H-26=37\Rightarrow H=63$.
\dots\ Wait, this seems to work!''}\\[1pt]
\hspace*{1em}$\downarrow$\\[1pt]
\emph{``Thus, the height of the cave is 63.''}}}
\end{minipage}\hfill
\begin{minipage}[t]{0.492\textwidth}
\fcolorbox{crose}{white}{\parbox[t]{\dimexpr\linewidth-2\fboxsep-2\fboxrule}{%
\textbf{\textcolor{crose}{Math specialist: answers 61, wrong}} \hfill 0 of 4 samples\\[3pt]
\emph{``$(y-x)^2+400=(y-x)^2+576$, which implies $400=576$, which is
impossible. Hmm, so this approach is invalid.''}\\[1pt]
\hspace*{1em}$\downarrow$\\[1pt]
\emph{``I think I have to concede and say the answer is 61.''}\\[1pt]
\hspace*{1em}$\downarrow$\\[1pt]
\emph{``The correct configuration involves the horizontal and vertical
distances being the components of the square's sides.''} (the same opening
returns in later steps)\\[1pt]
\hspace*{1em}$\downarrow$\\[1pt]
\emph{``Thus, the height of the cave is 61.''}}}
\end{minipage}

\vspace{5pt}
\fcolorbox{cink}{cice}{\parbox{\dimexpr\textwidth-2\fboxsep-2\fboxrule}{%
\raggedright\textbf{Task (MBPP+).} Write a function to check whether the
given number is undulating or not.
\texttt{assert is\_undulating(1212121) == True}}}

\vspace{4pt}
\begin{minipage}[t]{0.492\textwidth}
\fcolorbox{cnavy}{white}{\parbox[t]{\dimexpr\linewidth-2\fboxsep-2\fboxrule}{%
\textbf{\textcolor{cnavy}{\method: passes all tests}}\\[2pt]
{\ttfamily a, b = s[0], s[1]\\
if a == b:\\
\hspace*{2em}return False\\
for i in range(2, len(s)):\\
\hspace*{2em}if s[i] != s[i-2]:\\
\hspace*{4em}return False}}}
\end{minipage}\hfill
\begin{minipage}[t]{0.492\textwidth}
\fcolorbox{crose}{white}{\parbox[t]{\dimexpr\linewidth-2\fboxsep-2\fboxrule}{%
\textbf{\textcolor{crose}{Code specialist: fails in all four seeds}}\\[2pt]
{\ttfamily def is\_undulating(n):\\
\hspace*{2em}s = str(n)\\
\hspace*{2em}return s == s[::-1]}\\[3pt]
This seed tests for a palindrome and fails on 1991. The other three pass the
base tests and fail the additional ones.}}
\end{minipage}
\caption{\textbf{Trajectories side by side.} Key steps and lines, quoted from
one sampled answer of each model, on a geometry problem (top) and a code
task (bottom) that \method solves and the specialist of the domain does not.}
\label{fig:cases}
\end{figure}

The fused model is not only a compromise between its sources: on some
problems it succeeds where every sample of that domain's specialist fails.
Figure~\ref{fig:cases} sets two trajectories side by side on a geometry
problem of HMMT25. The fused model notices that the two diagonals of a
square share a midpoint, which settles the problem in a few lines, and it
reaches the answer in three of four samples. The mathematics specialist
assumes a relation between side lengths, runs into a contradiction, begins to
repeat one sentence, and ends with an unsupported answer; all four samples
are wrong. Its failure has the shape of the looping answers that
Section~\ref{sec:mechanism} finds among its training rollouts, and the fused
model, built without those answers, does not follow it there. On a code
task the fused model implements the rule of the task description and the
coding specialist, in every seed, does not (Appendix~\ref{app:cases}).

\subsection{Qualitative analysis}
\label{sec:mechanism}

\begin{figure}[t]
\centering
\begin{minipage}[t]{0.49\textwidth}
\vspace{0pt}\centering
\includegraphics[width=\linewidth]{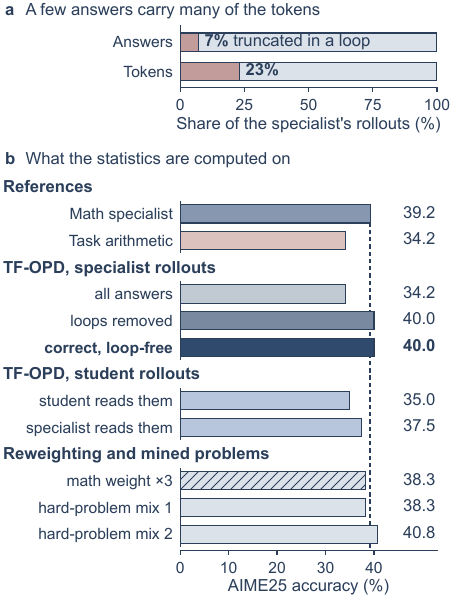}
\caption{\textbf{A few looping answers decide the hard problems.}
(a)~Answers that end in a repetition loop are a small share of the
mathematics specialist's answers and a large share of its tokens.
(b)~AIME25 accuracy of fused models by the answers their statistics use;
the dashed line is the mathematics specialist in float16, like them. The
final model of Table~\ref{tab:main} also keeps only verified code answers
(Appendix~\ref{app:ablations}).}
\label{fig:rollouts}
\end{minipage}\hfill
\begin{minipage}[t]{0.475\textwidth}
\vspace{0pt}\centering
\includegraphics[width=\linewidth]{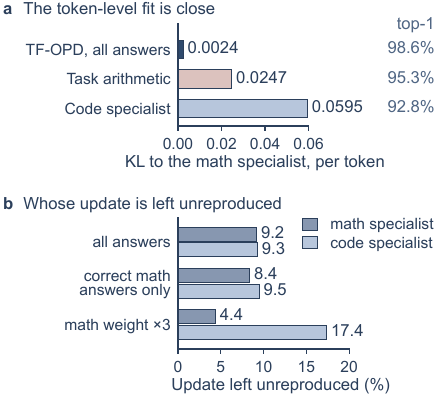}
\caption{\textbf{Following the specialist closely is not enough.} (a)~How
far each model is from the mathematics specialist when both read the
specialist's own correct solutions: KL divergence per token, and how often
the most likely next token agrees. (b)~Share of each specialist's update
that the fused layers fail to reproduce (Equation~\eqref{eq:diag}, median
over layers), for three choices of statistics.}
\label{fig:diag}
\end{minipage}
\end{figure}

\paragraph{A few answers never end.} Most of a specialist's answers are
ordinary reasoning that ends with a result. A few never end: the model starts
repeating itself and continues to the length limit. For the mathematics
specialist these are a small share of the answers but, being the longest, a
several times larger share of the tokens (Figure~\ref{fig:rollouts}a). Every
token counts equally in the statistics, so a fused model built from all
answers is asked, to a large extent, to behave as the specialist does inside
a loop.

\paragraph{The damage does not show token by token.} When both read the
specialist's own correct solutions, the fused model built from all answers
predicts almost the same next token as the specialist, far more closely than
task arithmetic does (Figure~\ref{fig:diag}a). Yet it scores the same as
task arithmetic on AIME25. A long derivation is decided at a few of its
steps, and an average over tokens does not see them.

\paragraph{Mathematics pays, code does not.} The repeated text concentrates
the mathematics statistics on a few directions, so each layer moves toward
the coding specialist wherever the two share inputs
(Proposition~\ref{prop:conflict}). The loss is confined to hard mathematics:
with all answers, the fused model still matches the coding specialist on both
code benchmarks and is slightly ahead of the mathematics specialist on
MATH-500 (Table~\ref{tab:abl-all}).

\subsection{Ablation study}
\label{sec:exp-selection}\label{sec:ablations}

\begin{table}[t]
\centering\small
\setlength{\tabcolsep}{8pt}
\begin{tabular}{@{}lccc@{}}
\toprule
Model & MATH-500 & HumanEval+ & MBPP+ \\
\midrule
\multicolumn{4}{@{}l}{\emph{No statistics}} \\
Task arithmetic & 92.0 & 81.7 & 69.3 \\
Low-rank sum & 91.6 & 82.9 & 68.8 \\
\midrule
\multicolumn{4}{@{}l}{\emph{\method, specialist rollouts}} \\
128 problems, 16k tokens & 92.0 & 84.8 & 68.5 \\
256 problems, 32k tokens & 93.2 & 87.2 & 70.1 \\
+ correct math answers only & 91.8 & 81.7 & 68.3 \\
+ verified code answers (final) & 90.8 & 84.1 & 67.5 \\
\midrule
\multicolumn{4}{@{}l}{\emph{\method, student rollouts}} \\
student reads its answers & 92.0 & 84.1 & 70.1 \\
+ correct math answers only & 92.4 & 84.8 & 69.0 \\
\bottomrule
\end{tabular}
\caption{\textbf{On the easier benchmarks every rule looks alike.} Accuracy
(\%), one sampled answer per problem. A row marked + adds one change to the
row above it.}
\label{tab:panel}
\end{table}

\paragraph{Which answers.} Figure~\ref{fig:rollouts}b varies only the answers
that enter the statistics. With every answer of the specialists, the fused
model is no better than task arithmetic on AIME25, and both fall short of the
mathematics specialist scored in the same format. Removing the looping
answers alone brings it level with the specialist and recovers every hard
problem that the model built from all answers loses. Keeping only correct
answers, which discards many more, changes nothing further: a finished but
wrong answer is still an example of the specialist reasoning normally.
Checking the code answers by running their tests, as the final model does,
changes code accuracy by no measurable amount (Table~\ref{tab:panel}).

\paragraph{Whose text.} On-policy distillation in the strict sense uses text
written by the student. We let a fused student answer the same problems and
compute the statistics either from the student reading its own answers or
from the mathematics specialist reading them, the closed-form analogue of
the trained objective. Neither improves on the specialist's own answers with
the looping ones removed (Figure~\ref{fig:rollouts}b).

\paragraph{Reweighting and mined problems.} Weighting mathematics three to
one in Equation~\eqref{eq:obj} recovers most of the loss, but at the coding
specialist's expense: the share of its update that the fused layers fail to
reproduce nearly doubles (Figure~\ref{fig:diag}b). Statistics from mined hard
problems give no clear gain. Selection removes the cause; the other remedies
compensate for it.

\paragraph{The merge rule.} On MATH-500 and the two code benchmarks, task
arithmetic, a low-rank variant of it and the closed form all keep both
specialists' gains and cannot be told apart (Table~\ref{tab:panel}). The
rules separate only on hard problems (Figure~\ref{fig:rollouts}b).

\paragraph{Number format.}\label{sec:exp-precision} Reinforcement learning
moves each weight by about one step of the 16-bit format the models are
released in. The closed-form update is not made of whole steps, and rounding
to that format measurably distorts it (Appendix~\ref{app:eval}), so fused
models are stored in 32-bit floats and evaluated in float16. The format
matters at the scale of our comparisons: the mathematics specialist scores
several points higher on AIME25 in its released format than in float16
(Table~\ref{tab:abl-aime}).

\subsection{Discussion}
\label{sec:discussion}

On-policy distillation is run after reinforcement learning to put several
specialists into one model. For specialists that share a base model, our
results suggest that the training in that step may not be needed. What the
trained student is asked to do, match each teacher on the text the models
generate, has an exact layer-wise solution, and that solution already gives a
model close to each teacher and more accurate on average than either. Adding
a specialist then costs its rollouts and one more term in each layer's sums;
nothing is retrained. Training remains the tool when the teachers differ in
architecture or size from the student.

\section{Conclusion}
\label{sec:conclusion}

Aggregating the capabilities of separately tuned specialists does not require
training-based on-policy distillation when they share a base model, and the
aggregate can be more accurate on average than any of its sources.
For a linear layer the distillation objective is least squares, its solution is
exact given the second moments of each teacher's inputs, and those are
obtained by letting the teachers read a few hundred of their own answers, at
the cost of forward passes. What those answers contain decides the hard
problems: removing a few looping answers, which carry a large share of the
tokens, lets the fused model match its mathematics specialist in the same
format.

\clearpage
\bibliographystyle{unsrtnat}
\bibliography{main}

@article{he2025justrl,
  author       = {Bingxiang He and
                  Zekai Qu and
                  Zeyuan Liu and
                  Yinghao Chen and
                  Yuxin Zuo and
                  Cheng Qian and
                  Kaiyan Zhang and
                  Weize Chen and
                  Chaojun Xiao and
                  Ganqu Cui and
                  Ning Ding and
                  Zhiyuan Liu},
  title        = {JustRL: Scaling a 1.5B {LLM} with a Simple {RL} Recipe},
  journal      = {CoRR},
  volume       = {abs/2512.16649},
  year         = {2025},
  url          = {https://doi.org/10.48550/arXiv.2512.16649},
  doi          = {10.48550/ARXIV.2512.16649},
  eprinttype   = {arXiv},
  eprint       = {2512.16649},
  bibsource    = {dblp computer science bibliography, https://dblp.org}
}

@article{shao2024deepseekmath,
  author       = {Zhihong Shao and
                  Peiyi Wang and
                  Qihao Zhu and
                  Runxin Xu and
                  Junxiao Song and
                  Mingchuan Zhang and
                  Y. K. Li and
                  Y. Wu and
                  Daya Guo},
  title        = {DeepSeekMath: Pushing the Limits of Mathematical Reasoning in Open
                  Language Models},
  journal      = {CoRR},
  volume       = {abs/2402.03300},
  year         = {2024},
  url          = {https://doi.org/10.48550/arXiv.2402.03300},
  doi          = {10.48550/ARXIV.2402.03300},
  eprinttype   = {arXiv},
  eprint       = {2402.03300},
  bibsource    = {dblp computer science bibliography, https://dblp.org}
}

@inproceedings{yu2025dapo,
  author       = {Qiying Yu and
                  Zheng Zhang and
                  Ruofei Zhu and
                  Yufeng Yuan and
                  Xiaochen Zuo and
                  Yu Yue and
                  Weinan Dai and
                  Tiantian Fan and
                  Gaohong Liu and
                  Juncai Liu and
                  Lingjun Liu and
                  Xin Liu and
                  Haibin Lin and
                  Zhiqi Lin and
                  Bole Ma and
                  Guangming Sheng and
                  Yuxuan Tong and
                  Chi Zhang and
                  Mofan Zhang and
                  Ru Zhang and
                  Wang Zhang and
                  Hang Zhu and
                  Jinhua Zhu and
                  Jiaze Chen and
                  Jiangjie Chen and
                  Chengyi Wang and
                  Hongli Yu and
                  Yuxuan Song and
                  Xiangpeng Wei and
                  Hao Zhou and
                  Jingjing Liu and
                  Wei{-}Ying Ma and
                  Ya{-}Qin Zhang and
                  Lin Yan and
                  Yonghui Wu and
                  Mingxuan Wang},
  editor       = {Danielle Belgrave and
                  Cheng Zhang and
                  Laura N. Montoya and
                  Hsuan{-}Tien Lin and
                  Razvan Pascanu and
                  Piotr Koniusz and
                  Marzyeh Ghassemi and
                  Nancy Chen and
                  Iv{\'{a}}n Vladimir Meza Ru{\'{\i}}z and
                  Arturo Loaiza{-}Bonilla},
  title        = {{DAPO:} An Open-Source {LLM} Reinforcement Learning System at Scale},
  booktitle    = {Advances in Neural Information Processing Systems 38: Annual Conference
                  on Neural Information Processing Systems 2025, NeurIPS 2025, San Diego,
                  CA, USA, December 2-7, 2025 / Mexico City, Mexico, November 30 - December
                  5, 2025},
  year         = {2025},
  url          = {http://papers.nips.cc/paper\_files/paper/2025/hash/a4277440d50f1f15d2cb4c14f7e0c0d2-Abstract-Conference.html},
  bibsource    = {dblp computer science bibliography, https://dblp.org}
}

@inproceedings{wortsman2022soups,
  author       = {Mitchell Wortsman and
                  Gabriel Ilharco and
                  Samir Yitzhak Gadre and
                  Rebecca Roelofs and
                  Raphael Gontijo Lopes and
                  Ari S. Morcos and
                  Hongseok Namkoong and
                  Ali Farhadi and
                  Yair Carmon and
                  Simon Kornblith and
                  Ludwig Schmidt},
  editor       = {Kamalika Chaudhuri and
                  Stefanie Jegelka and
                  Le Song and
                  Csaba Szepesv{\'{a}}ri and
                  Gang Niu and
                  Sivan Sabato},
  title        = {Model soups: averaging weights of multiple fine-tuned models improves
                  accuracy without increasing inference time},
  booktitle    = {International Conference on Machine Learning, {ICML} 2022, 17-23 July
                  2022, Baltimore, Maryland, {USA}},
  series       = {Proceedings of Machine Learning Research},
  volume       = {162},
  pages        = {23965--23998},
  publisher    = {{PMLR}},
  year         = {2022},
  url          = {https://proceedings.mlr.press/v162/wortsman22a.html},
  bibsource    = {dblp computer science bibliography, https://dblp.org}
}

@inproceedings{ilharco2023task,
  author       = {Gabriel Ilharco and
                  Marco T{\'{u}}lio Ribeiro and
                  Mitchell Wortsman and
                  Ludwig Schmidt and
                  Hannaneh Hajishirzi and
                  Ali Farhadi},
  title        = {Editing models with task arithmetic},
  booktitle    = {The Eleventh International Conference on Learning Representations,
                  {ICLR} 2023, Kigali, Rwanda, May 1-5, 2023},
  publisher    = {OpenReview.net},
  year         = {2023},
  url          = {https://openreview.net/forum?id=6t0Kwf8-jrj},
  bibsource    = {dblp computer science bibliography, https://dblp.org}
}

@inproceedings{yadav2023ties,
  author       = {Prateek Yadav and
                  Derek Tam and
                  Leshem Choshen and
                  Colin A. Raffel and
                  Mohit Bansal},
  editor       = {Alice Oh and
                  Tristan Naumann and
                  Amir Globerson and
                  Kate Saenko and
                  Moritz Hardt and
                  Sergey Levine},
  title        = {TIES-Merging: Resolving Interference When Merging Models},
  booktitle    = {Advances in Neural Information Processing Systems 36: Annual Conference
                  on Neural Information Processing Systems 2023, NeurIPS 2023, New Orleans,
                  LA, USA, December 10 - 16, 2023},
  year         = {2023},
  url          = {http://papers.nips.cc/paper\_files/paper/2023/hash/1644c9af28ab7916874f6fd6228a9bcf-Abstract-Conference.html},
  bibsource    = {dblp computer science bibliography, https://dblp.org}
}

@inproceedings{yu2024dare,
  author       = {Le Yu and
                  Bowen Yu and
                  Haiyang Yu and
                  Fei Huang and
                  Yongbin Li},
  editor       = {Ruslan Salakhutdinov and
                  Zico Kolter and
                  Katherine A. Heller and
                  Adrian Weller and
                  Nuria Oliver and
                  Jonathan Scarlett and
                  Felix Berkenkamp},
  title        = {Language Models are Super Mario: Absorbing Abilities from Homologous
                  Models as a Free Lunch},
  booktitle    = {Forty-first International Conference on Machine Learning, {ICML} 2024,
                  Vienna, Austria, July 21-27, 2024},
  series       = {Proceedings of Machine Learning Research},
  volume       = {235},
  pages        = {57755--57775},
  publisher    = {{PMLR} / OpenReview.net},
  year         = {2024},
  url          = {https://proceedings.mlr.press/v235/yu24p.html},
  bibsource    = {dblp computer science bibliography, https://dblp.org}
}

@article{deepseek2025r1,
  author       = {DeepSeek{-}AI},
  title        = {DeepSeek-R1: Incentivizing Reasoning Capability in LLMs via Reinforcement
                  Learning},
  journal      = {CoRR},
  volume       = {abs/2501.12948},
  year         = {2025},
  url          = {https://doi.org/10.48550/arXiv.2501.12948},
  doi          = {10.48550/ARXIV.2501.12948},
  eprinttype   = {arXiv},
  eprint       = {2501.12948},
  bibsource    = {dblp computer science bibliography, https://dblp.org}
}

@inproceedings{hendrycks2021math,
  author       = {Dan Hendrycks and
                  Collin Burns and
                  Saurav Kadavath and
                  Akul Arora and
                  Steven Basart and
                  Eric Tang and
                  Dawn Song and
                  Jacob Steinhardt},
  editor       = {Joaquin Vanschoren and
                  Sai{-}Kit Yeung},
  title        = {Measuring Mathematical Problem Solving With the {MATH} Dataset},
  booktitle    = {Proceedings of the Neural Information Processing Systems Track on
                  Datasets and Benchmarks 1, NeurIPS Datasets and Benchmarks 2021, December
                  2021, virtual},
  year         = {2021},
  url          = {https://datasets-benchmarks-proceedings.neurips.cc/paper/2021/hash/be83ab3ecd0db773eb2dc1b0a17836a1-Abstract-round2.html},
  bibsource    = {dblp computer science bibliography, https://dblp.org}
}

@inproceedings{balunovic2025matharena,
  author       = {Mislav Balunovic and
                  Jasper Dekoninck and
                  Ivo Petrov and
                  Nikola Jovanovic and
                  Martin T. Vechev},
  editor       = {Danielle Belgrave and
                  Cheng Zhang and
                  Laura N. Montoya and
                  Hsuan{-}Tien Lin and
                  Razvan Pascanu and
                  Piotr Koniusz and
                  Marzyeh Ghassemi and
                  Nancy Chen and
                  Iv{\'{a}}n Vladimir Meza Ru{\'{\i}}z and
                  Arturo Loaiza{-}Bonilla},
  title        = {MathArena: Evaluating LLMs on Uncontaminated Math Competitions},
  booktitle    = {Advances in Neural Information Processing Systems 38: Annual Conference
                  on Neural Information Processing Systems 2025, NeurIPS 2025, San Diego,
                  CA, USA, December 2-7, 2025 / Mexico City, Mexico, November 30 - December
                  5, 2025},
  year         = {2025},
  url          = {http://papers.nips.cc/paper\_files/paper/2025/hash/1d27c01ebd3e3aebe226b44fc970d803-Abstract-Datasets\_and\_Benchmarks\_Track.html},
  bibsource    = {dblp computer science bibliography, https://dblp.org}
}

@inproceedings{liu2025prorl,
  author       = {Mingjie Liu and
                  Shizhe Diao and
                  Ximing Lu and
                  Jian Hu and
                  Xin Dong and
                  Yejin Choi and
                  Jan Kautz and
                  Yi Dong},
  editor       = {Danielle Belgrave and
                  Cheng Zhang and
                  Laura N. Montoya and
                  Hsuan{-}Tien Lin and
                  Razvan Pascanu and
                  Piotr Koniusz and
                  Marzyeh Ghassemi and
                  Nancy Chen and
                  Iv{\'{a}}n Vladimir Meza Ru{\'{\i}}z and
                  Arturo Loaiza{-}Bonilla},
  title        = {ProRL: Prolonged Reinforcement Learning Expands Reasoning Boundaries
                  in Large Language Models},
  booktitle    = {Advances in Neural Information Processing Systems 38: Annual Conference
                  on Neural Information Processing Systems 2025, NeurIPS 2025, San Diego,
                  CA, USA, December 2-7, 2025 / Mexico City, Mexico, November 30 - December
                  5, 2025},
  year         = {2025},
  url          = {http://papers.nips.cc/paper\_files/paper/2025/hash/1a22b912945fb7c0bdd079e792b31b6f-Abstract-Conference.html},
  bibsource    = {dblp computer science bibliography, https://dblp.org}
}

@inproceedings{matena2021fisher,
  author       = {Michael Matena and
                  Colin Raffel},
  editor       = {Sanmi Koyejo and
                  S. Mohamed and
                  A. Agarwal and
                  Danielle Belgrave and
                  K. Cho and
                  A. Oh},
  title        = {Merging Models with Fisher-Weighted Averaging},
  booktitle    = {Advances in Neural Information Processing Systems 35: Annual Conference
                  on Neural Information Processing Systems 2022, NeurIPS 2022, New Orleans,
                  LA, USA, November 28 - December 9, 2022},
  year         = {2022},
  url          = {http://papers.nips.cc/paper\_files/paper/2022/hash/70c26937fbf3d4600b69a129031b66ec-Abstract-Conference.html},
  bibsource    = {dblp computer science bibliography, https://dblp.org}
}

@inproceedings{gargiulo2024tsv,
  author       = {Antonio Andrea Gargiulo and
                  Donato Crisostomi and
                  Maria Sofia Bucarelli and
                  Simone Scardapane and
                  Fabrizio Silvestri and
                  Emanuele Rodol{\`{a}}},
  title        = {Task Singular Vectors: Reducing Task Interference in Model Merging},
  booktitle    = {{IEEE/CVF} Conference on Computer Vision and Pattern Recognition,
                  {CVPR} 2025, Nashville, TN, USA, June 11-15, 2025},
  pages        = {18695--18705},
  publisher    = {Computer Vision Foundation / {IEEE}},
  year         = {2025},
  url          = {https://openaccess.thecvf.com/content/CVPR2025/html/Gargiulo\_Task\_Singular\_Vectors\_Reducing\_Task\_Interference\_in\_Model\_Merging\_CVPR\_2025\_paper.html},
  doi          = {10.1109/CVPR52734.2025.01742},
  bibsource    = {dblp computer science bibliography, https://dblp.org}
}

@article{yang2026doesscalingreinforcementlearning,
  title={Does Scaling Reinforcement Learning Really Require More Training?},
  author={Yang, Bangji and Fan, Jiajun and Ma, Hongba and Guo, Ruihan and Liu, Ge},
  journal={arXiv preprint arXiv:2610.01133},
  year={2026}
}

@inproceedings{marczak2025isotropic,
  author       = {Daniel Marczak and
                  Simone Magistri and
                  Sebastian Cygert and
                  Bartlomiej Twardowski and
                  Andrew D. Bagdanov and
                  Joost van de Weijer},
  editor       = {Aarti Singh and
                  Maryam Fazel and
                  Daniel Hsu and
                  Simon Lacoste{-}Julien and
                  Felix Berkenkamp and
                  Tegan Maharaj and
                  Kiri Wagstaff and
                  Jerry Zhu},
  title        = {No Task Left Behind: Isotropic Model Merging with Common and Task-Specific
                  Subspaces},
  booktitle    = {Forty-second International Conference on Machine Learning, {ICML}
                  2025, Vancouver, BC, Canada, July 13-19, 2025},
  series       = {Proceedings of Machine Learning Research},
  volume       = {267},
  publisher    = {{PMLR} / OpenReview.net},
  year         = {2025},
  url          = {https://proceedings.mlr.press/v267/marczak25a.html},
  bibsource    = {dblp computer science bibliography, https://dblp.org}
}

@article{ma2026s3spectralnullspaceswap,
  title={$S^3$: Spectral Null-Space Swap Makes Reasoning Models Efficient},
  author={Ma, Hongbo and Cao, Sansheng and Fan, Jiajun and Yang, Bangji and Liu, Ge},
  journal={arXiv preprint arXiv:2609.37976},
  year={2026}
}

@article{olmo2025olmo3,
  author       = {Allyson Ettinger and
                  Amanda Bertsch and
                  Bailey Kuehl and
                  David Graham and
                  David Heineman and
                  Dirk Groeneveld and
                  Faeze Brahman and
                  Finbarr Timbers and
                  Hamish Ivison and
                  Jacob Morrison and
                  Jake Poznanski and
                  Kyle Lo and
                  Luca Soldaini and
                  Matt Jordan and
                  Mayee F. Chen and
                  Michael Noukhovitch and
                  Nathan Lambert and
                  Pete Walsh and
                  Pradeep Dasigi and
                  Robert Berry and
                  Saumya Malik and
                  Saurabh Shah and
                  Scott Geng and
                  Shane Arora and
                  Shashank Gupta and
                  Taira Anderson and
                  Teng Xiao and
                  Tyler Murray and
                  Tyler Romero and
                  Victoria Graf and
                  Akari Asai and
                  Akshita Bhagia and
                  Alexander Wettig and
                  Alisa Liu and
                  Aman Rangapur and
                  Chloe Anastasiades and
                  Costa Huang and
                  Dustin Schwenk and
                  Harsh Trivedi and
                  Ian Magnusson and
                  Jaron Lochner and
                  Jiacheng Liu and
                  Lester James V. Miranda and
                  Maarten Sap and
                  Malia Morgan and
                  Michael Schmitz and
                  Michal Guerquin and
                  Michael Wilson and
                  Regan Huff and
                  Ronan Le Bras and
                  Rui Xin and
                  Rulin Shao and
                  Sam Skjonsberg and
                  Shannon Zejiang Shen and
                  Shuyue Stella Li and
                  Tucker Wilde and
                  Valentina Pyatkin and
                  William Merrill and
                  Yapei Chang and
                  Yuling Gu and
                  Zhiyuan Zeng and
                  Ashish Sabharwal and
                  Luke Zettlemoyer and
                  Pang Wei Koh and
                  Ali Farhadi and
                  Noah A. Smith and
                  Hannaneh Hajishirzi},
  title        = {Olmo 3},
  journal      = {CoRR},
  volume       = {abs/2512.13961},
  year         = {2025},
  url          = {https://doi.org/10.48550/arXiv.2512.13961},
  doi          = {10.48550/ARXIV.2512.13961},
  eprinttype   = {arXiv},
  eprint       = {2512.13961},
  bibsource    = {dblp computer science bibliography, https://dblp.org}
}

@inproceedings{liu2023evalplus,
  author       = {Jiawei Liu and
                  Chunqiu Steven Xia and
                  Yuyao Wang and
                  Lingming Zhang},
  editor       = {Alice Oh and
                  Tristan Naumann and
                  Amir Globerson and
                  Kate Saenko and
                  Moritz Hardt and
                  Sergey Levine},
  title        = {Is Your Code Generated by ChatGPT Really Correct? Rigorous Evaluation
                  of Large Language Models for Code Generation},
  booktitle    = {Advances in Neural Information Processing Systems 36: Annual Conference
                  on Neural Information Processing Systems 2023, NeurIPS 2023, New Orleans,
                  LA, USA, December 10 - 16, 2023},
  year         = {2023},
  url          = {http://papers.nips.cc/paper\_files/paper/2023/hash/43e9d647ccd3e4b7b5baab53f0368686-Abstract-Conference.html},
  bibsource    = {dblp computer science bibliography, https://dblp.org}
}

@article{sun2026resmerge,
  author       = {Yandu Sun and
                  ZhiYan Hou and
                  Haokai Ma and
                  Yuheng Jia and
                  Junfeng Fang and
                  Haiyun Guo and
                  Hongyan An and
                  Weizhen Wang and
                  Jinqiao Wang},
  title        = {ResMerge: Residual-based Spectral Merging of Large Language Models},
  journal      = {CoRR},
  volume       = {abs/2606.02252},
  year         = {2026},
  url          = {https://doi.org/10.48550/arXiv.2606.02252},
  doi          = {10.48550/ARXIV.2606.02252},
  eprinttype   = {arXiv},
  eprint       = {2606.02252},
  bibsource    = {dblp computer science bibliography, https://dblp.org}
}

@article{hinton2015distilling,
  author       = {Geoffrey E. Hinton and
                  Oriol Vinyals and
                  Jeffrey Dean},
  title        = {Distilling the Knowledge in a Neural Network},
  journal      = {CoRR},
  volume       = {abs/1503.02531},
  year         = {2015},
  url          = {https://doi.org/10.48550/arXiv.1503.02531},
  doi          = {10.48550/ARXIV.1503.02531},
  eprinttype   = {arXiv},
  eprint       = {1503.02531},
  bibsource    = {dblp computer science bibliography, https://dblp.org}
}

@inproceedings{kim2016sequence,
  author       = {Yoon Kim and
                  Alexander M. Rush},
  editor       = {Jian Su and
                  Xavier Carreras and
                  Kevin Duh},
  title        = {Sequence-Level Knowledge Distillation},
  booktitle    = {Proceedings of the 2016 Conference on Empirical Methods in Natural
                  Language Processing, {EMNLP} 2016, Austin, Texas, USA, November 1-4,
                  2016},
  pages        = {1317--1327},
  publisher    = {The Association for Computational Linguistics},
  year         = {2016},
  url          = {https://doi.org/10.18653/v1/d16-1139},
  doi          = {10.18653/V1/D16-1139},
  bibsource    = {dblp computer science bibliography, https://dblp.org}
}

@inproceedings{ross2011dagger,
  author       = {St{\'{e}}phane Ross and
                  Geoffrey J. Gordon and
                  Drew Bagnell},
  editor       = {Geoffrey J. Gordon and
                  David B. Dunson and
                  Miroslav Dud{\'{\i}}k},
  title        = {A Reduction of Imitation Learning and Structured Prediction to No-Regret
                  Online Learning},
  booktitle    = {Proceedings of the Fourteenth International Conference on Artificial
                  Intelligence and Statistics, {AISTATS} 2011, Fort Lauderdale, USA,
                  April 11-13, 2011},
  series       = {{JMLR} Proceedings},
  volume       = {15},
  pages        = {627--635},
  publisher    = {JMLR.org},
  year         = {2011},
  url          = {http://proceedings.mlr.press/v15/ross11a/ross11a.pdf},
  bibsource    = {dblp computer science bibliography, https://dblp.org}
}

@inproceedings{agarwal2024gkd,
  author       = {Rishabh Agarwal and
                  Nino Vieillard and
                  Yongchao Zhou and
                  Piotr Stanczyk and
                  Sabela Ramos Garea and
                  Matthieu Geist and
                  Olivier Bachem},
  title        = {On-Policy Distillation of Language Models: Learning from Self-Generated
                  Mistakes},
  booktitle    = {The Twelfth International Conference on Learning Representations,
                  {ICLR} 2024, Vienna, Austria, May 7-11, 2024},
  publisher    = {OpenReview.net},
  year         = {2024},
  url          = {https://openreview.net/forum?id=3zKtaqxLhW},
  bibsource    = {dblp computer science bibliography, https://dblp.org}
}

@inproceedings{gu2024minillm,
  author       = {Yuxian Gu and
                  Li Dong and
                  Furu Wei and
                  Minlie Huang},
  title        = {MiniLLM: Knowledge Distillation of Large Language Models},
  booktitle    = {The Twelfth International Conference on Learning Representations,
                  {ICLR} 2024, Vienna, Austria, May 7-11, 2024},
  publisher    = {OpenReview.net},
  year         = {2024},
  url          = {https://openreview.net/forum?id=5h0qf7IBZZ},
  bibsource    = {dblp computer science bibliography, https://dblp.org}
}

@inproceedings{wan2024fusellm,
  author       = {Fanqi Wan and
                  Xinting Huang and
                  Deng Cai and
                  Xiaojun Quan and
                  Wei Bi and
                  Shuming Shi},
  title        = {Knowledge Fusion of Large Language Models},
  booktitle    = {The Twelfth International Conference on Learning Representations,
                  {ICLR} 2024, Vienna, Austria, May 7-11, 2024},
  publisher    = {OpenReview.net},
  year         = {2024},
  url          = {https://openreview.net/forum?id=jiDsk12qcz},
  bibsource    = {dblp computer science bibliography, https://dblp.org}
}

@inproceedings{zhao2026opsd,
title={Self-Distilled Reasoner: On-Policy Self-Distillation for Large Language Models},
author={Siyan Zhao and Zhihui Xie and Mengchen Liu and Jing Huang and Guan Pang and Feiyu Chen and Aditya Grover},
booktitle={Forty-third International Conference on Machine Learning},
year={2026},
url={https://openreview.net/forum?id=Jpxfof0EaS}
}

@article{chen2021codex,
  author       = {Mark Chen and
                  Jerry Tworek and
                  Heewoo Jun and
                  Qiming Yuan and
                  Henrique Pond{\'{e}} de Oliveira Pinto and
                  Jared Kaplan and
                  Harri Edwards and
                  Yuri Burda and
                  Nicholas Joseph and
                  Greg Brockman and
                  Alex Ray and
                  Raul Puri and
                  Gretchen Krueger and
                  Michael Petrov and
                  Heidy Khlaaf and
                  Girish Sastry and
                  Pamela Mishkin and
                  Brooke Chan and
                  Scott Gray and
                  Nick Ryder and
                  Mikhail Pavlov and
                  Alethea Power and
                  Lukasz Kaiser and
                  Mohammad Bavarian and
                  Clemens Winter and
                  Philippe Tillet and
                  Felipe Petroski Such and
                  Dave Cummings and
                  Matthias Plappert and
                  Fotios Chantzis and
                  Elizabeth Barnes and
                  Ariel Herbert{-}Voss and
                  William Hebgen Guss and
                  Alex Nichol and
                  Alex Paino and
                  Nikolas Tezak and
                  Jie Tang and
                  Igor Babuschkin and
                  Suchir Balaji and
                  Shantanu Jain and
                  William Saunders and
                  Christopher Hesse and
                  Andrew N. Carr and
                  Jan Leike and
                  Joshua Achiam and
                  Vedant Misra and
                  Evan Morikawa and
                  Alec Radford and
                  Matthew Knight and
                  Miles Brundage and
                  Mira Murati and
                  Katie Mayer and
                  Peter Welinder and
                  Bob McGrew and
                  Dario Amodei and
                  Sam McCandlish and
                  Ilya Sutskever and
                  Wojciech Zaremba},
  title        = {Evaluating Large Language Models Trained on Code},
  journal      = {CoRR},
  volume       = {abs/2107.03374},
  year         = {2021},
  url          = {https://doi.org/10.48550/arXiv.2107.03374},
  doi          = {10.48550/ARXIV.2107.03374},
  eprinttype   = {arXiv},
  eprint       = {2107.03374},
  bibsource    = {dblp computer science bibliography, https://dblp.org}
}

@article{austin2021mbpp,
  author       = {Jacob Austin and
                  Augustus Odena and
                  Maxwell I. Nye and
                  Maarten Bosma and
                  Henryk Michalewski and
                  David Dohan and
                  Ellen Jiang and
                  Carrie J. Cai and
                  Michael Terry and
                  Quoc V. Le and
                  Charles Sutton},
  title        = {Program Synthesis with Large Language Models},
  journal      = {CoRR},
  volume       = {abs/2108.07732},
  year         = {2021},
  url          = {https://doi.org/10.48550/arXiv.2108.07732},
  doi          = {10.48550/ARXIV.2108.07732},
  eprinttype   = {arXiv},
  eprint       = {2108.07732},
  bibsource    = {dblp computer science bibliography, https://dblp.org}
}

@article{luo2025deepscaler,
  title={Deepscaler: Surpassing o1-preview with a 1.5 b model by scaling rl},
  author={Luo, Michael and Tan, Sijun and Wong, Justin and Shi, Xiaoxiang and Tang, William Y and Roongta, Manan and Cai, Colin and Luo, Jeffrey and Li, Li Erran and Popa, Raluca Ada and others},
  journal={Notion Blog},
  volume={3},
  number={5},
  year={2025}
}

@article{luo2025deepcoder,
  title={Deepcoder: A fully open-source 14b coder at o3-mini level},
  author={Luo, Michael and Tan, Sijun and Huang, Roy and Patel, Ameen and Ariyak, Alpay and Wu, Qingyang and Shi, Xiaoxiang and Xin, Rachel and Cai, Colin and Weber, Maurice and others},
  journal={Notion Blog},
  volume={1},
  year={2025}
}

@inproceedings{jin2023regmean,
  author       = {Xisen Jin and
                  Xiang Ren and
                  Daniel Preotiuc{-}Pietro and
                  Pengxiang Cheng},
  title        = {Dataless Knowledge Fusion by Merging Weights of Language Models},
  booktitle    = {The Eleventh International Conference on Learning Representations,
                  {ICLR} 2023, Kigali, Rwanda, May 1-5, 2023},
  publisher    = {OpenReview.net},
  year         = {2023},
  url          = {https://openreview.net/forum?id=FCnohuR6AnM},
  bibsource    = {dblp computer science bibliography, https://dblp.org}
}

@article{hou2026whatmatters,
  author       = {Zhinan Hou and
                  Jiaqi Zhang and
                  Xunliang Cai and
                  Keyou You},
  title        = {What Matters in On-Policy Distillation? {A} Perspective on Data Efficiency
                  and Data Selection},
  journal      = {CoRR},
  volume       = {abs/2609.05198},
  year         = {2026},
  url          = {https://doi.org/10.48550/arXiv.2609.05198},
  doi          = {10.48550/ARXIV.2609.05198},
  eprinttype   = {arXiv},
  eprint       = {2609.05198},
  bibsource    = {dblp computer science bibliography, https://dblp.org}
}

@article{xu2026paced,
  title={PACED: Distillation and on-policy self-distillation at the frontier of student competence},
  author={Xu, Yuanda and Sang, Hejian and Zhou, Zhengze and He, Ran and Wang, Zhipeng},
  journal={arXiv preprint arXiv:2603.11178},
  year={2026}
}

@article{hou2026uniopd,
  author       = {Wenjin Hou and
                  Shangpin Peng and
                  Weinong Wang and
                  Zheng Ruan and
                  Yue Zhang and
                  Zhenglin Zhou and
                  Mingqi Gao and
                  Yifei Chen and
                  Kaiqi Wang and
                  Hongming Yang and
                  Chengquan Zhang and
                  Zhuotao Tian and
                  Han Hu and
                  Yi Yang and
                  Fei Wu and
                  Hehe Fan},
  title        = {Uni-OPD: Unifying On-Policy Distillation with a Dual-Perspective Recipe},
  journal      = {CoRR},
  volume       = {abs/2605.03677},
  year         = {2026},
  url          = {https://doi.org/10.48550/arXiv.2605.03677},
  doi          = {10.48550/ARXIV.2605.03677},
  eprinttype   = {arXiv},
  eprint       = {2605.03677},
  bibsource    = {dblp computer science bibliography, https://dblp.org}
}

@inproceedings{he2024olympiadbench,
  author       = {Chaoqun He and
                  Renjie Luo and
                  Yuzhuo Bai and
                  Shengding Hu and
                  Zhen Leng Thai and
                  Junhao Shen and
                  Jinyi Hu and
                  Xu Han and
                  Yujie Huang and
                  Yuxiang Zhang and
                  Jie Liu and
                  Lei Qi and
                  Zhiyuan Liu and
                  Maosong Sun},
  editor       = {Lun{-}Wei Ku and
                  Andre Martins and
                  Vivek Srikumar},
  title        = {OlympiadBench: {A} Challenging Benchmark for Promoting {AGI} with
                  Olympiad-Level Bilingual Multimodal Scientific Problems},
  booktitle    = {Proceedings of the 62nd Annual Meeting of the Association for Computational
                  Linguistics (Volume 1: Long Papers), {ACL} 2024, Bangkok, Thailand,
                  August 11-16, 2024},
  pages        = {3828--3850},
  publisher    = {Association for Computational Linguistics},
  year         = {2024},
  url          = {https://doi.org/10.18653/v1/2024.acl-long.211},
  doi          = {10.18653/V1/2024.ACL-LONG.211},
  bibsource    = {dblp computer science bibliography, https://dblp.org}
}

@inproceedings{liang2025boosting,
  title={Boosting Multi-Domain Fine-Tuning of Large Language Models through Evolving Interactions between Samples},
  author={Liang, Xize and Yang, Lin and Wang, Jie and Lu, Yiyang and Wu, Runyu and Chen, Hanzhu and Hao, Jianye},
  booktitle={International Conference on Machine Learning},
  pages={37427--37441},
  year={2025},
  organization={PMLR}
}

@inproceedings{liang2026boosting,
  title={Boosting Multi-Domain Reasoning of LLMs via Curvature-Guided Policy Optimization},
  author={Liang, Xize and Yang, Lin and Wang, Jie and Liu, Rui and Lu, Yang and Zeng, Jinliang and Chen, Hanzhu and Li, Dong and Hao, Jianye},
  booktitle={The Fourteenth International Conference on Learning Representations},
  year={2026}
}

@article{liu2026socraticpopolicyoptimizationinteractive,
  title={Socraticpo: Policy optimization via interactive guidance},
  author={Liu, Zirui and Pan, Tingyue and Ouyang, Jie and Liu, Qi and Wang, Xianquan and Liu, Jiayu and Li, Qingchuan and Sha, Jing and Huang, Zhenya and Wang, Shijin and others},
  journal={arXiv preprint arXiv:2606.09887},
  year={2026}
}

@inproceedings{liu2025elo,
  title={am-ELO: A Stable Framework for Arena-based LLM Evaluation},
  author={Liu, Zirui and Li, Jiatong and Zhuang, Yan and Liu, Qi and Shen, Shuanghong and Ouyang, Jie and Cheng, Mingyue and Wang, Shijin},
  booktitle={International Conference on Machine Learning},
  pages={38857--38868},
  year={2025},
  organization={PMLR}
}

@inproceedings{lin2026models,
  title={When Models Learn to Ask Why: Adaptive Causal Reasoning for Trustworthy Medical Vision-Language Models},
  author={Lin, Jianxin and Zhu, Chunzheng and Kneuertz, Peter J and Bai, Yunfei and Xue, Yuan},
  booktitle={Proceedings of the IEEE/CVF Conference on Computer Vision and Pattern Recognition},
  pages={5556--5568},
  year={2026}
}

@inproceedings{zhu2026medsynapsevbridgingvisualperception,
  title={Medsynapse-v: Bridging visual perception and clinical intuition via latent memory evolution},
  author={Zhu, Chunzheng and Wang, Yijun and Zeng, Jiaqi and Jiang, Junyu and Lin, Jianxin},
  booktitle={European Conference on Computer Vision},
  pages={559--577},
  year={2026},
  organization={Springer}
}

@article{huang2025adaptive,
  title={Adaptive Sample Scheduling for Direct Preference Optimization},
  author={Huang, Zixuan and Ban, Yikun and Fu, Lean and Li, Xiaojie and Dai, Zhongxiang and Li, Jianxin and Wang, Deqing},
  journal={arXiv preprint arXiv:2506.17252},
  year={2025}
}

@article{Huang2026RealTimeAR,
  title={Real-time aligned reward model beyond semantics},
  author={Huang, Zixuan and Xia, Xin and Ren, Yuxi and Zheng, Jianbin and Xiao, Xuefeng and Xie, Hongyan and Huaqiu, Li and Liang, Songshi and Dai, Zhongxiang and Zhuang, Fuzhen and others},
  journal={arXiv preprint arXiv:2601.22664},
  year={2026}
}

@inproceedings{NEURIPS2023_32285dd1,
 author = {Cai, Xin-Qiang and Zhang, Pushi and Zhao, Li and Bian, Jiang and Sugiyama, Masashi and Llorens, Ashley},
 booktitle = {Advances in Neural Information Processing Systems},
 doi = {10.52202/075280-0686},
 editor = {A. Oh and T. Naumann and A. Globerson and K. Saenko and M. Hardt and S. Levine},
 pages = {15593--15613},
 publisher = {Curran Associates, Inc.},
 title = {Distributional Pareto-Optimal Multi-Objective Reinforcement Learning},
 url = {https://proceedings.neurips.cc/paper_files/paper/2023/file/32285dd184dbfc33cb2d1f0db53c23c5-Paper-Conference.pdf},
 volume = {36},
 year = {2023}
}

\clearpage
\appendix
\raggedbottom

\section{Derivation and properties}
\label{app:derivation}

\paragraph{Closed form.} With $G_d=\mathbb{E}_{x\sim\mu_d}[xx^{\!\top}]$, the
objective of Equation~\eqref{eq:obj} is
\begin{equation*}
J(W)=\sum_d w_d\,\tr\!\big[(W-W_d)\,G_d\,(W-W_d)^{\!\top}\big],
\end{equation*}
which is convex and quadratic in $W$, with gradient
$\nabla J=2\sum_d w_d\,(W-W_d)\,G_d$. Setting it to zero gives
\begin{equation*}
W\sum_d w_dG_d=\sum_d w_dW_dG_d .
\end{equation*}
Substituting $W_d=\Wb+\dd_d$, the term $\Wb\sum_dw_dG_d$ appears on both
sides, so $W=\Wb+\dd_\star$ with $\dd_\star\sum_d\Gt_d=\sum_d\dd_d\Gt_d$, which
is Equation~\eqref{eq:solve}. The solution is the same whether one solves for
the weight or for the update.

\paragraph{Shrinkage as a penalty.} Replacing $G_d$ by
$\alpha G_d+(1-\alpha)\,\mathrm{diag}(G_d)$ replaces teacher $d$'s term of $J$
by
\begin{equation*}
\alpha\,\mathbb{E}\big\|(W-W_d)x\big\|_2^2
+(1-\alpha)\sum_i(G_d)_{ii}\,\big\|(W-W_d)_{:,i}\big\|_2^2 .
\end{equation*}
The second term penalizes the deviation of each input column of $W$ from the
teacher's, in proportion to that coordinate's second moment. Since
$\mathrm{diag}(G_d)$ is positive definite whenever every input coordinate is
active, $\sum_d\Gt_d$ is invertible for $\alpha<1$.

\paragraph{Proposition~\ref{prop:conflict}(i): identical statistics give the
average.} If $G_d=G$ for all $d$, write
$\Gt=\alpha G+(1-\alpha)\,\mathrm{diag}(G)$. Then
\begin{equation*}
\dd_\star=\Big(\sum_dw_d\dd_d\Big)\Gt\,\Big(\sum_dw_d\,\Gt\Big)^{-1}
=\frac{\sum_dw_d\dd_d}{\sum_dw_d}.
\end{equation*}
With equal weights this is the mean of the updates, not their sum: \method
reduces to weight averaging, not to task arithmetic, when the teachers cannot
be told apart by their inputs.

\paragraph{Proposition~\ref{prop:conflict}(ii): a dominant teacher is
recovered.} As $w_d/w_e\to0$ for every $d\ne e$, we have
$\sum_d\dd_d\Gt_d\to\dd_e\Gt_e$ and $\sum_d\Gt_d\to\Gt_e$, so
$\dd_\star\to\dd_e$.

\paragraph{Proposition~\ref{prop:conflict}(iii): orthogonal inputs are
reproduced exactly.} Take $\alpha=1$, so that $\Gt_d=w_dG_d$, and suppose the
ranges of $G_1,\dots,G_K$ are mutually orthogonal. Then
$(\sum_d\Gt_d)^{+}=\sum_d\Gt_d^{+}$ and $\Gt_d\Gt_e^{+}=0$ for $d\ne e$, so the
minimum-norm solution
\begin{equation*}
\dd_\star=\Big(\sum_d\dd_d\Gt_d\Big)\Big(\sum_d\Gt_d\Big)^{+}
\end{equation*}
satisfies $\dd_\star x=\dd_e\Gt_e\Gt_e^{+}x=\dd_ex$ for every $x$ in the range of
$G_e$. Each teacher is reproduced without error on its own inputs, and
$J(W_\star)=0$. Interference between teachers is confined to the input
directions they share.

\paragraph{Task arithmetic in the same terms.} Task arithmetic uses
$\dd=\sum_d\dd_d$, whose error on teacher $e$'s inputs is
$\sum_{d\ne e}\dd_dx$. Its objective value is therefore
\begin{equation*}
J=\sum_ew_e\,\tr\!\Big[\Big(\sum_{d\ne e}\dd_d\Big)G_e\Big(\sum_{d\ne e}\dd_d\Big)^{\!\top}\Big],
\end{equation*}
which is nonzero unless the other teachers' updates sum to zero on teacher
$e$'s inputs. \method attains the minimum of the same $J$, up to the shrinkage
of Equation~\eqref{eq:shrink}.

\paragraph{The diagnostic.} The numerator of Equation~\eqref{eq:diag} is
teacher $d$'s term of $J$, with the shrinkage of Equation~\eqref{eq:shrink},
at the solution, and the denominator is its value at $\dd_\star=0$, the base
model. Hence $\varepsilon_d=0$ when the teacher is reproduced on its inputs,
and $\varepsilon_d=1$ when the fused layer does no better than the base layer.

\FloatBarrier
\section{Implementation and evaluation protocol}
\label{app:eval}

Algorithm~\ref{alg:main} summarizes \method; the paragraphs below give the
details of each step and of the evaluation.

\begin{algorithm}[h]
\caption{\method}
\label{alg:main}
\begin{algorithmic}[1]
\Require base $\theta_{\mathrm{b}}$; teachers $\theta_1,\dots,\theta_K$;
training problems $\mathcal{P}_d$ per domain; weights $w_d$; $\alpha$
\For{$d=1$ \textbf{to} $K$} \Comment{forward only}
  \State $\mathcal{R}_d\gets$ answers of $\theta_d$ to $\mathcal{P}_d$
  \State $\mathcal{S}_d\gets\{r\in\mathcal{R}_d:\ r$ correct, finished, no loop$\}$
  \State run $\theta_d$ on $\mathcal{S}_d$ with teacher forcing
  \State $G_{d,g}\gets X_g^{\!\top}X_g/n_d$ for each layer group $g$
\EndFor
\For{each group $g$ and layer $\ell\in g$} \Comment{solve}
  \State $\Gt_{d}\gets w_d[\alpha G_{d,g}+(1-\alpha)\mathrm{diag}(G_{d,g})]$
  \State $W^\ell_\star\gets\Wb^\ell+\big(\sum_d\dd^\ell_d\Gt_{d}\big)\big(\sum_d\Gt_{d}\big)^{-1}$
\EndFor
\State other tensors: $\Wb+\sum_d\dd_d$
\State \Return $\theta_\star$, stored in 32-bit floats
\end{algorithmic}
\end{algorithm}

\paragraph{Models.} All three Olmo~3 models have 32 layers, hidden size 4096
and a vocabulary of 100{,}278, and are released under Apache~2.0.
\begin{itemize}
\item Base: {\small\texttt{allenai/Olmo-3-1025-7B}}
\item Mathematics: {\small\texttt{allenai/Olmo-3.1-7B-RL-Zero-Math}}
\item Code: {\small\texttt{allenai/Olmo-3.1-7B-RL-Zero-Code}}
\end{itemize}
The solve warns when a teacher's weights differ from the base model's by more
than half their norm.

\paragraph{Rollouts.} Prompts are formatted with the chat template each
teacher was trained with. Mathematics: 256 problems from DAPO-Math-17k, one
answer each, temperature 0.7, top-$p$ 0.9, at most 31{,}744 new tokens. Code:
256 problems from the DeepCoder training set, prefixed with the EvalPlus
instruction, one answer each, temperature 1.0, top-$p$ 1.0, at most 32{,}768
new tokens. Generation uses vLLM. Both problem sets are training sets; no
evaluation problem is used.

\paragraph{Selection.} An answer is \emph{finished} if generation stopped
before the length limit. It is a \emph{loop} if the last 20{,}000 characters
have a DEFLATE compression ratio below 0.05. Of the 256 mathematics answers,
19 are truncated, and all 19 are repetition loops; two of them state an answer
before they loop, are graded correct, and are removed with the rest.
Mathematics answers are graded as in evaluation, and selection keeps the 175
that are correct, finished and not loops. No code answer is truncated or a
loop. Code answers are graded by running the last Python block of the answer
on the tests that come with the problem, in a sandbox with limits on memory,
time and output, on the 241 problems that read standard input; 92 pass every
test, and these are kept. The reference solutions of the dataset pass the same
grader on 212 of the 241 problems.

\paragraph{Statistics.} The teacher is loaded in bfloat16 and reads each kept
sequence, truncated to 36{,}864 tokens, with teacher forcing. A forward
pre-hook on each group of linear layers accumulates $X^{\!\top}X$ in 32-bit
floats over all positions of prompt and answer, and the sum is divided by the
number of tokens. Groups are found automatically: a short sequence is passed
through the model once, and linear layers of a block that receive the
identical tensor are assigned to one group. In Olmo~3 this gives four groups
per block (query, key and value; attention output; gate and up; down), and
the final hidden states provide the statistics of the output head.

\paragraph{Solve.} $\alpha=0.9$ and $w_d=1$. Each system is solved in 64-bit
floats with one linear solve per layer, and the result is stored in 32-bit
floats. The output head is solved from the final-hidden-state statistics when
it is not tied to the input embedding. Embeddings, normalization gains and
biases use task arithmetic. The solve also reports the cosine between the
fused update and the same update after the fused weights are rounded to
bfloat16, which is about 0.96 for our fused models.

\paragraph{Evaluation.} Mathematics benchmarks use temperature 0.7, top-$p$
0.9 and at most 31{,}744 new tokens. AIME24, AIME25 and HMMT25 have thirty
problems each; four answers are sampled per problem and the mean correctness
over all answers is reported. MATH-500 and OlympiadBench (674 problems) use
one sampled answer per problem.
An answer is correct if the content of its last
\texttt{\textbackslash boxed\{\}} expression, or of its final ``Answer:''
line, matches the reference after normalization or is symbolically equal to
it; answers these rules reject are re-checked by CompassVerifier-3B. Code
benchmarks use temperature 1.0, top-$p$ 1.0 and at most 32{,}768 new tokens,
with one sampled answer per problem; the last Python code block that defines
the required entry point is extracted and scored by EvalPlus on the base and
additional tests, and an answer that reaches the length limit counts as
failed. Released checkpoints are evaluated in bfloat16 and fused models in
float16. For the study of Section~\ref{sec:exp-selection} the mathematics
specialist is evaluated in float16 as well (Table~\ref{tab:abl-aime}).

\paragraph{Compute.} All runs use 48\,GB GPUs, one per process.
Table~\ref{tab:cost} lists the cost of each step. Generating the rollouts
dominates, the teacher-forced pass and the solve are short, and no step
computes a gradient or keeps optimizer state.

\begin{table}[t]
\centering\small
\setlength{\tabcolsep}{8pt}
\begin{tabular}{@{}llr@{}}
\toprule
Step & Unit & Cost \\
\midrule
Rollouts & 64 problems & 1 GPU-hour \\
Statistics & 2.6M tokens & 40 min \\
Solve & all layers & 25 min \\
\midrule
Total & two specialists & $\approx$10 GPU-hours \\
\midrule
AIME25 evaluation & one model & 2.5--3 GPU-hours \\
\bottomrule
\end{tabular}
\caption{\textbf{Cost of \method on Olmo~3 7B, on 48\,GB GPUs.} Every step is
forward computation or linear algebra, and none computes a gradient. The total
is derived from the rows above for two specialists with 256 training problems
each; evaluating one model on AIME25 is given for scale.}
\label{tab:cost}
\end{table}

\FloatBarrier
\section{Ablations and diagnostics}
\label{app:ablations}

All fused models of this appendix use the solve of
Section~\ref{sec:method} with $\alpha=0.9$, are stored in float32 and
evaluated in float16, and differ only in the statistics they are given.
Differences between two models are paired over problems, with 95\% intervals
from 10{,}000 bootstrap resamples.

\paragraph{Weight-only rules.} Task arithmetic adds both updates to the base
weights. The low-rank variant takes the singular value decomposition of each
update, treats singular values above 1.05 times the Marchenko--Pastur edge as
the low-rank part, sums the low-rank parts of the two specialists and
averages the rest.

\paragraph{Amount of rollouts.} Table~\ref{tab:panel} compares two amounts of
rollouts: 128 training problems per domain with a limit of 16k new tokens,
and 256 other problems per domain at the limits of evaluation. Unless stated
otherwise, closed-form models use the latter (2.6 million answer tokens for
mathematics with all answers, 1.2 million after selection).

\paragraph{Student rollouts.} The student is the closed-form model built from
128 problems per domain with a limit of 16k new tokens. It answers the same
256 problems under the evaluation settings. In the first variant the student
reads its own answers, so that the statistics are those of the student's
activations; a version of it that keeps only the student's correct
mathematics answers is also reported (Table~\ref{tab:panel}). In the second
variant the problems are restricted to the 165 that the teacher solves within
12k tokens, the student's answers to them are kept after removing loops, and
the mathematics teacher reads them. This is the closed-form counterpart of the
trained on-policy objective, in which the teacher supervises text written by
the student \citep{hou2026whatmatters,xu2026paced,hou2026uniopd}.

\paragraph{Domain weights.} The mathematics Gram matrices of the
all-answers statistics are multiplied by three.

\paragraph{Mined hard problems.} From 2{,}000 further DAPO-Math-17k
problems we take 400 on which a 1.5B reasoning model solves at most two of
four samples. The mathematics teacher and a fused student (all answers)
answer each once, with a limit of 16k new tokens. Problems solved by the
teacher and not by the student, and problems solved by both, give four sets
of statistics, all read by the mathematics teacher: the student's and the
teacher's answers on each class, with loops in student answers cut at their
onset. Mix~1 weights the student's answers on the first class, the teacher's
answers on the first class and the student's answers on the second by 0.4,
0.4 and 0.2. Mix~2 weights the selected teacher statistics of the final model
and the teacher's answers on the two classes by 0.5, 0.25 and 0.25. Teacher
and student solve about the same number of these problems, so the classes
reflect sampling luck more than a capability gap. The code statistics are
unchanged.

\begin{table}[t]
\centering\small
\begin{minipage}[t]{0.44\textwidth}
\vspace{0pt}\centering
\setlength{\tabcolsep}{5pt}
\begin{tabular}{@{}lc@{}}
\toprule
Model & AIME25 \\
\midrule
Mathematics specialist, bfloat16 & 43.3 \\
Mathematics specialist, float16  & 39.2 \\
Task arithmetic                  & 34.2 \\
\midrule
\multicolumn{2}{@{}l}{\emph{\method, specialist rollouts}} \\
All answers                     & 34.2 \\
Looping answers removed         & 40.0 \\
Correct, finished, loop-free    & 40.0 \\
\midrule
\multicolumn{2}{@{}l}{\emph{\method, student rollouts}} \\
Student reads its answers       & 35.0 \\
Specialist reads the student's answers & 37.5 \\
\midrule
\multicolumn{2}{@{}l}{\emph{Reweighting and mined problems}} \\
All answers, math weight $\times$3 & 38.3 \\
Hard-problem mix 1              & 38.3 \\
Hard-problem mix 2              & 40.8 \\
\bottomrule
\end{tabular}
\caption{\textbf{AIME25 by what the statistics are computed on.} Accuracy
(\%), four sampled answers per problem. The mathematics specialist is shown
in both evaluation formats; fused models are evaluated in float16. Task
arithmetic is stored in bfloat16, which represents the sum of the two
updates almost exactly.}
\label{tab:abl-aime}
\end{minipage}\hfill
\begin{minipage}[t]{0.51\textwidth}
\vspace{0pt}\centering
\setlength{\tabcolsep}{4pt}
\begin{tabular}{@{}lrc@{}}
\toprule
Comparison & Diff. & 95\% interval \\
\midrule
Specialist: float16 $-$ bfloat16        & $-4.2$ & $[-10.8,\,+1.7]$ \\
All answers $-$ specialist (bfloat16)   & $-9.2$ & $[-15.8,\,-3.3]$ \\
Correct, loop-free $-$ all answers      & $+5.8$ & $[-1.7,\,+13.3]$ \\
Student reads $-$ all answers           & $+0.8$ & $[-4.2,\,+5.8]$ \\
Specialist reads $-$ all answers        & $+3.3$ & $[-3.3,\,+10.0]$ \\
Math weight $\times$3 $-$ all answers   & $+4.2$ & $[-2.5,\,+10.8]$ \\
Mix 1 $-$ correct, loop-free            & $-1.7$ & $[-6.7,\,+3.3]$ \\
Mix 2 $-$ correct, loop-free            & $+0.8$ & $[-4.2,\,+5.8]$ \\
\bottomrule
\end{tabular}
\caption{\textbf{Paired differences on AIME25.} Difference in accuracy
(points) between two models of Table~\ref{tab:abl-aime}, with its 95\%
bootstrap interval.}
\label{tab:abl-ci}

\vspace{14pt}
\setlength{\tabcolsep}{3.2pt}
\begin{tabular}{@{}lccc@{}}
\toprule
Model & MATH-500 & HumanEval+ & MBPP+ \\
\midrule
On-domain specialist & 91.4 & 83.2 & 69.6 \\
\method, all answers & 92.9 & 83.8 & 70.9 \\
\bottomrule
\end{tabular}
\caption{\textbf{The fused model built from all answers, away from hard
mathematics.} Accuracy (\%) averaged over four sampled answers per problem,
against the specialist of each benchmark's domain.}
\label{tab:abl-all}
\end{minipage}
\end{table}

\paragraph{Results.} Table~\ref{tab:abl-aime} gives the values of
Figure~\ref{fig:rollouts}b and Table~\ref{tab:abl-ci} the paired differences.
The final model of Table~\ref{tab:main} uses the mathematics statistics of
the correct, loop-free row together with the verified code statistics, so it
is a different model from that row. With thirty problems the intervals are
wide: the one difference that excludes zero is the loss of the all-answers
model to the mathematics specialist. On MATH-500 the all-answers model is
ahead of the mathematics specialist by a margin whose interval excludes zero,
and on the code benchmarks it is level with the coding specialist
(Table~\ref{tab:abl-all}), which is why the loss on hard problems does not
show on the saturated benchmarks.

\begin{table}[t]
\centering\small
\setlength{\tabcolsep}{8pt}
\begin{tabular}{@{}lccc@{}}
\toprule
 & KL per & Top-1 & Residual \\
Model & token & (\%) & (\%) \\
\midrule
\method, all answers & 0.0024 & 98.6 & 0.9 $\to$ 3.6 \\
Task arithmetic & 0.0247 & 95.3 & 1.2 $\to$ 10.4 \\
Code specialist & 0.0595 & 92.8 & 1.7 $\to$ 18.6 \\
\bottomrule
\end{tabular}
\caption{\textbf{Teacher-forced agreement.} Agreement with the mathematics
specialist under teacher forcing along its own correct solutions to AIME25
and MATH-500 problems: KL divergence per token, top-1 agreement, and relative
deviation of the residual stream at the first and at the last layer.}
\label{tab:diag}
\end{table}

\paragraph{Teacher-forced agreement.} For Figure~\ref{fig:diag}a and
Table~\ref{tab:diag} each model reads, with teacher forcing, 51 of the
mathematics specialist's own correct solutions to AIME25 problems and 64 to
MATH-500 problems, and its next-token distribution and residual stream are
compared with the specialist's at every position. The agreement does not
degrade with the position in the answer, and it is the same on both
benchmarks. The student-rollout model has the same agreement as the
all-answers model.

\paragraph{Unreproduced share.} Table~\ref{tab:eps} gives the diagnostic of
Equation~\eqref{eq:diag}. It is largest for the attention query and key
projections and smallest for the attention output projection, larger in the
early layers than in the late ones, and small for the output head. Keeping
only correct, loop-free mathematics answers shifts it slightly in favor of
mathematics; weighting mathematics three to one halves it for mathematics and
nearly doubles it for code.

\begin{table}[t]
\centering\small
\begin{minipage}[t]{0.40\textwidth}
\vspace{0pt}\centering
\setlength{\tabcolsep}{5pt}
\begin{tabular*}{\linewidth}{@{\extracolsep{\fill}}lc@{}}
\toprule
Matrix & Share (\%) \\
\midrule
Attn. query & 13.0 \\
Attn. key & 15.0 \\
Attn. value & 8.5 \\
Attn. output & 5.0 \\
MLP gate & 11.5 \\
MLP up & 10.7 \\
MLP down & 7.8 \\
Output head, mathematics & 1.9 \\
Output head, code & 6.6 \\
\bottomrule
\end{tabular*}
\end{minipage}\hspace{0.06\textwidth}%
\begin{minipage}[t]{0.40\textwidth}
\vspace{0pt}\centering
\setlength{\tabcolsep}{5pt}
\begin{tabular*}{\linewidth}{@{\extracolsep{\fill}}lcc@{}}
\toprule
Statistics & Math & Code \\
\midrule
All answers & 9.2 & 9.3 \\
Correct math answers only & 8.4 & 9.5 \\
Math weight $\times$3 & 4.4 & 17.4 \\
\bottomrule
\end{tabular*}
\end{minipage}
\caption{\textbf{Unreproduced share of a teacher's update
(Equation~\eqref{eq:diag}).} Median over layers, in percent. Left: by matrix
type, for the all-answers statistics. Right: by teacher, for three sets of
statistics.}
\label{tab:eps}
\end{table}

\FloatBarrier
\section{Case studies}
\label{app:cases}

The case studies are problems on which the fused model of
Table~\ref{tab:main} succeeds while the specialist of that domain fails in
every sample. So that a specialist's failure is not a single unlucky sample,
the specialist is sampled several times: four samples per model on the
competition benchmarks, and one sample of the fused model against four seeds
of the coding specialist on the code benchmarks. Problems that meet this
criterion include HMMT25 problems 21 and 26 and AIME25 problem 25 for
mathematics, and HumanEval/126, HumanEval/93, Mbpp/92, Mbpp/244, Mbpp/278,
Mbpp/563 and Mbpp/301 for code. We describe three of them.

\begin{figure}[h]
\centering
\includegraphics[width=0.5\textwidth]{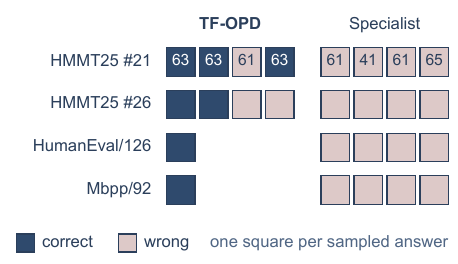}
\caption{\textbf{Problems that \method solves and the specialist of the
domain does not.} One square per sampled answer: four per model on HMMT25,
and one for \method against four seeds of the coding specialist on the code
benchmarks. Numbers are the final answers given on HMMT25 problem~21, whose
reference answer is 63.}
\label{fig:cases-grid}
\end{figure}

\paragraph{HMMT25, problem 21.} \emph{In a two-dimensional cave with a
parallel floor and ceiling, two stalactites of lengths 16 and 36 hang
perpendicularly from the ceiling, while two stalagmites of heights 25 and 49
grow perpendicularly from the ground. If the tips of these four structures
form the vertices of a square in some order, compute the height of the
cave.} The answer is 63. The fused model answers 63 in three of its four
samples (Figures~\ref{fig:cases} and~\ref{fig:cases-grid}). It uses the fact
that the diagonals of a square share a midpoint: pairing the two stalactite
tips as one diagonal and the two stalagmite tips as the other, the midpoint
heights agree when $H-26=37$. The mathematics specialist is wrong in all four
samples. In the sample quoted in Figure~\ref{fig:cases}, it assumes a relation
between side lengths that leads to a contradiction, repeats one sentence about
``the correct configuration'', and ends with an unsupported answer.

\paragraph{HumanEval/126.} The task is to decide whether a list is sorted in
ascending order, returning false as well when a number occurs more than
twice. The fused model checks the order and then counts runs of equal
numbers, and passes the base and the additional tests. The coding specialist
mishandles the rule about repeated numbers in all four seeds and fails the
base tests.

\paragraph{Mbpp/92.} The task is to decide whether a number is undulating,
that is, whether two different digits alternate. The fused model requires
the first two digits to differ and every later digit to equal the digit two
places before it, and passes all tests. The coding specialist fails in all
four seeds: one tests for a palindrome and fails the base tests, and three
pass the base tests and fail the additional edge cases.

\end{document}